\documentclass{article}
\usepackage{graphicx}
\usepackage{amssymb}
\usepackage{amsmath}
\usepackage{subfigure} 
\usepackage{algorithm}
\usepackage{algorithmic}
\usepackage{hyperref}
\usepackage{bbm}

\usepackage[accepted]{icml2013}
\usepackage{natbib}
\setcitestyle{square, numbers}
\icmltitlerunning{Generalized Latent Variable Recovery for Generative Adversarial Networks}
\begin{document}
\twocolumn[
\icmltitle{Generalized Latent Variable Recovery \protect\\ for Generative Adversarial Networks}
\icmlauthor{Nicholas Egan}{negan@mit.edu}
\icmlauthor{Jeffrey Zhang}{jeffreyz@mit.edu}
\icmlauthor{Kevin Shen}{kevshen@mit.edu}
\icmladdress{Massachusetts Institute of Technology, Department of Electrical Engineering and Computer Science}
\vskip 0.3in
]
\small

\begin{abstract} 
The Generator of a Generative Adversarial Network (GAN) is trained to transform latent vectors drawn from a prior distribution into realistic looking photos. These latent vectors have been shown to encode information about the content of their corresponding images. Projecting input images onto the latent space of a GAN is non-trivial, but previous work has successfully performed this task for latent spaces with a uniform prior. We extend these techniques to latent spaces with a Gaussian prior, and demonstrate our technique's effectiveness.
\end{abstract} 

\section{Introduction}
Generative Adversarial Networks (GANs) are a class of generative neural network models that can be used to synthetically produce photo-realistic looking images. In the framework developed by Goodfellow et al. \cite{Goodfellow14}, we aim to train two models simultaneously using an adversarial process: a generative model $G$ (called the ``Generator") representing the data distribution, and a discriminative model $D$ (called the ``Discriminator") that estimates the probability that a given input came from the training data rather that the generative model $G$. More specifically, our ultimate goal is to learn the generator distribution $p_{g}$ over the data space $\mathbf{x}$. To do this, we will first define $p_{z}(z)$ to be the prior distribution on the input noise and $G(z; \theta_g)$ (the generative model) to be a map from the generator parameters $\theta_{g}$ and input noise $z$ to the data space. We also define another function $D(x; \theta_{d})$ (the discriminator) which takes in a sample and outputs a scalar value, representing the probability that the sample came from the data rather than $p_{g}$.

While GANs are extremely effective models when it comes to evaluating the images they produce, they are notoriously difficult to train, as the global optimum of the loss function is a saddle point as opposed to a minimum \cite{Creswell16}. As a result, many techniques have been developed recently to stabilize GAN training. One of these, proposed by White \cite{White16} is to use a Gaussian prior $p_z(z) = \mathcal{N}(0,I)$ as opposed to a uniform prior $p_z(z) = \mathcal{U}(0,1)$, which was previously the standard.

The latent space of a GAN has been shown to function well as a feature representation for images of a particular domain. Zhu et al. \cite{Zhu16} demonstrated that ``Euclidean distance in the latent space often corresponds to a perceptually meaningful visual similarity," and Radford et al. \cite{Radford16} demonstrated the usefulness of GANs for unsupervised representation learning. Thus the ability to reliably extract the latent representation of a given input image is quite useful. Earlier work in the space of latent variable recovery has focused on latent spaces with uniform priors. Techniques by Lipton et al. \cite{Lipton17} can recover latent variables with near perfect accuracy when the latent space has a uniform prior, but no techniques in the literature have demonstrated the same degree of precision when the latent space is Gaussian.

We attempt to solve this problem using techniques of probabilistic resampling. In this paper, we start by giving a high level overview of GANs and DCGANs. Next, we describe the problem of latent variable recovery in GANs. Afterwards, we present our experiments, results, and analysis. Lastly, we look at some potential applications of the methodology.

\section{Generative Adversarial Networks}
\subsection{Basic GAN}
$G(z; \theta_{g})$ and $D(x; \theta_{d})$ are typically modeled by two feedforward neural networks, since we can train these easily using backpropogation. In training, we want to maximize $D$'s probability of assigning the correct label to actual data points and samples from $G$. At the same time, we want to train $G$ to minimize $\log(1 - D(G(z)))$. The overall adversarial relationship can be summarized by the following minimax equation:
\begin{equation*}
\begin{split}
\min_G \max_D V(D,G) &= \mathbb{E}_{x \sim p_{\text{data}}(x)}[\log(D(x))] \\
&+ \mathbb{E}_{z \sim p_{z}(z)}[\log(1-D(G(z)))]
\end{split}
\end{equation*}
Because we are using differentiable functions to estimate our discriminator and generator, it is natural to then use stochastic gradient descent updates to train the parameters of the two functions. For example, the updates for the Discriminator and the Generator on a mini-batch of $m$ noise samples $\{ z^{(1)}, ..., z^{(m)}\}$ drawn from the noise prior $p_g(z)$ and a mini-batch of $m$ examples $\{x^{(1)}, ..., x^{(m)}\}$ from the underlying data distribution $p_{\text{data}}(x)$ is as follows:
\begin{gather*}
	\nabla_{\theta_d}\frac{1}{m}\sum_{i=1}^m \left[ \log D(x^{(i)}) + \log (1 - D(G(z^{(i)})))\right] \\
    \nabla_{\theta_g}\frac{1}{m}\sum_{i=1}^m \log (1 - D(G(z^{(i)})))
\end{gather*}
In the seminal paper on GAN's, Goodfellow et al. \cite{Goodfellow14} proved that an algorithm using the gradient updates above is guaranteed to reach the global optimum, if at each iteration $i$, the discriminator is allowed to reach it's optimum given the current generator $G^{(i)}$. Additionally, it was also shown that the global optimum for the generator distribution $p_g$ is the underlying distribution of the data $p_{\text{data}}$. To see this, first note that for any fixed generator $G$, the optimal discriminator $D$ is 
\begin{align*}
    D^*_g(x) = \frac{p_\text{data}(x)}{p_\text{data}(x) + p_g(x)}
\end{align*}
With this in mind, we can rewrite our minimax function as 
\begin{align*}
    C(G) &= \max_DV(G,D) \\
    &= \mathbb{E}_{x \sim p_{\text{data}}(x)}\left[\log\left(\frac{p_\text{data}(x)}{p_\text{data}(x) + p_g(x)}\right)\right] \\
&+ \mathbb{E}_{z \sim p_{z}(z)}\left[\log\left(\frac{p_g(x)}{p_\text{data}(x) + p_g(x)}\right)\right]
\end{align*}
The above virtual training criterion $C(G)$ is minimized when $p_g = p_{\text{data}}$ and $D^*_g(x) = \frac{1}{2}$ and has value $-\log(4)$ at its minimum. In addition, another way of seeing that the optimum occurs when $p_g = p_{\text{data}}$, is rewriting the virtual training criterion in terms of Kullback-Leibler divergence.
\begin{align*}
    C(G) = &-\log(4) + \text{KL}\left( p_\text{data}||\frac{ p_\text{data} + p_g}{2}\right) \\
    &+ \text{KL}\left( p_g||\frac{ p_\text{data} + p_g}{2}\right) 
\end{align*}
Thus, the closer our two distributions are to $\frac{ p_\text{data} + p_g}{2}$, the closer to the optimum we are, and we achieve the optimum when $p_g = p_\text{data} = \frac{ p_\text{data} + p_g}{2}$.

\subsection{DCGAN}
In this next section, we will discuss a special type of GANs called deep convolutional generative adversarial networks (DCGANs), which are commonly used for image generation. As the name suggests, the models are essentially our classic GAN models that use CNNs to represent both the discriminator and generator distributions. Historically, this approach has been unsuccessful and unstable to train. However, Radford et al. \cite{Radford16} recently showed that with some moderate constraints to the model architecture, we can achieve stable training. These new GANs are known as DCGANs. 

We will now discuss the architecture constraints specified by Radford et al. As previously mentioned, in DCGANs, the discriminator and generator distributions are both modeled by CNNs. The first constraint introduced is the use of fully convolutional networks (CNNs with no pooling layers and only strided convolutions). Fractionally-strided convolutional layers (transposed convolution) are used for the Generator, while strided convolutional layers are used for the Discriminator. The second constraint is the elimination of fully connected layers on top of convolutional features. The last constraint is the use of batch normalization, which normalizes the input to each unit to have mean 0 and unit variance. In addition to the above constraints, Radford et al. suggested using ReLU activation in the generator in all layers except for the output, which should use the tanh function. Additionally, the use of leaky ReLU in the discriminator was also suggested. A diagram of the DCGAN architecture can be found in Figure 1.
\begin{figure}[h]
\centering
  \begin{minipage}[h]{0.40\textwidth}
  \centering
    \includegraphics[width=\textwidth]{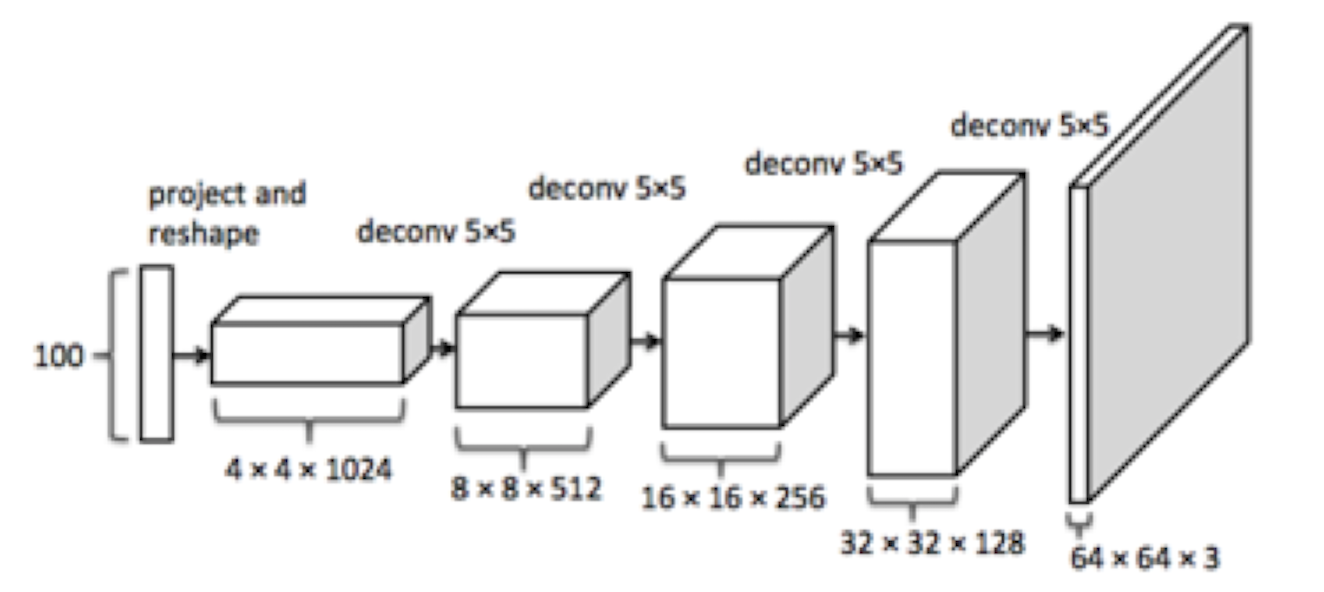}
    \includegraphics[width=\textwidth]{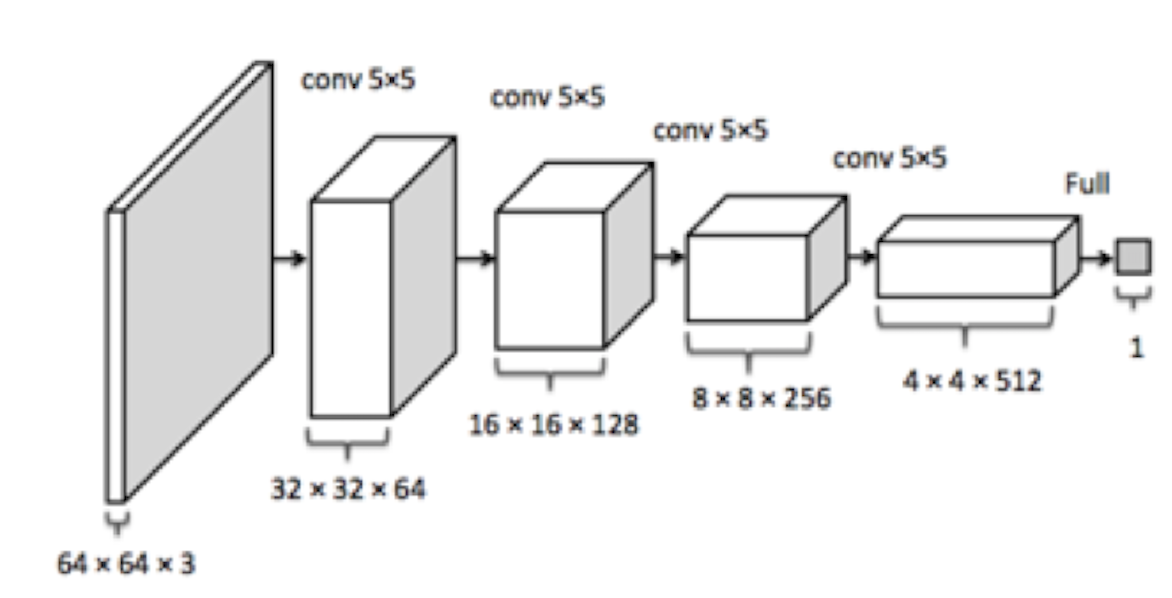}
    \caption{DCGAN Architecture - Generator (top) and Discriminator (bottom)}
  \end{minipage}
\end{figure}

\section{Latent Variable Recovery}
\subsection{Previous Work}
It is a common setup in generative models to take a latent vector $z$ as input and transform it into $x$, a realistic looking image. Based on the work of Kingma et al. \cite{Kingma14}, a Variational Autoencoder (VAE) is a generative model that is trained by taking an input image $x$, transforming it into a latent vector $z$ using an encoder network, and then transforming $z$ into $x'$ using a decoder network. Thus for a VAE, encoding an image into a latent vector or decoding a latent vector into an image is explicit, as networks are trained for
these tasks.

GANs also use a latent vector $z$ as input to the generative network, but this latent vector is drawn
from a prior distribution as opposed to being extracted from a training image. Thus the task of projecting
an input image onto the latent space of a GAN is non-trivial. Directly inverting the generator network is computationally expensive and sometimes impossible, because it is usually a multi-layered non-linear model.

Dumouli et al.\cite{Dumoulin14} and Donahue et al.\cite{Donahue17} perform latent
vector recovery by changing the GAN setup to include three networks: the discriminator, the generator,
and the inference network, where the inference network is trained alongside the other two networks to 
learn the task of mapping $x$ to $z$. This technique has mixed success, as recovered latent vectors
often fail to preserve key information about the content of their corresponding images. Additionally,
adding a third network requires additional parameters to be trained, and prevents latent variable 
reconstruction from being performed on pre-trained networks.

An alternative approach is to start with a guess $z' \sim p_z(z)$ and use gradient descent to move the image
generated from $z$ closer to the original image $x$. This technique was first mentioned by Zhu et al. \cite{Zhu16} as a step in a larger process to recover latent vectors, but shown by Creswell et al.
\cite{Creswell16} to perform fairly well as an effective technique in isolation with batching. Lipton et al. \cite{Lipton17} showed that when the prior $p_z(z)$ is a uniform distribution, latent vectors could be reconstructed with near perfect accuracy using a technique they call
stochastic clipping.

\subsection{Our Methods}
We attempt to extend the techniques of Lipton et al. \cite{Lipton17} to GANs with a Gaussian prior. The setup is as
follows: Given a pretrained GAN with generator $G(z)$, we wish to invert an image $x$ into a latent 
vector $z$ that approximates the original latent vector used to generate $x$. We start by initializing
$z^{(0)}$ to a random vector drawn from $p_z(z)$, which in our case is a multivariate unit normal
distribution. Loss is computed as:
$$ L(z, x) = ||x - G(z)||_2^2$$
$z$ is optimized using gradient descent, making updates of the form $$ z^{(t+1)} \leftarrow z^{(t)} - \eta^{(t)} \nabla_{z} L(z^{(t)},x) $$ where the learning rate $\eta^{(t)}$ is adjusted over time using the Adam optimizer. This optimization problem is non-convex, and using gradient descent alone often leads to components of the $z$ vector getting stuck in local minima during unlucky initializations.

The insight of Lipton et al. \cite{Lipton17} was that if the prior distribution $p_z(z)$ is a uniform distribution $\mathcal{U}(a,b)$ for each component of $z$, then $P(z_i < a) = 0$ and $P(z_i > b) = 0$ for all $z_i \in z$. Thus if any $z_i$ goes beyond the bounds set by the prior distribution during gradient descent, it should be resampled.

When $p_z(z)$ is a normal distribution however, there is no $z_i \in \mathbb{R}$ such that $P(z_i) = 0$.
We propose three approximate criteria to decide when to resample $z_i$ despite having non-zero probability
in order to drive our final $z'$ to a more probable result. We define $R_i$ as the probability that we
want to resample component $z_i$ of $z$, and give our formulations for $R_i$ below. We also define
$R_{i}^{(t)}$ as the probability that we want to resample $z_i$ at timestep $t$. What we are trying to
prevent is initializations where a component $z_i$ quickly becomes stuck in a local minimum. If we
expect $E$ iterations of our algorithm, then we can calculate the
probability that we want to resample a $z_i$ at any individual timestep $t$ as:
$$P(R_i^{(t)} | z_i, t) = P(R_i^{(t)} | z_i) = 1 - (1 - P(R_i | z_i))^{1/E} $$
Thus the full algorithm for reconstructing a latent vector is shown in \textbf{Algorithm 1}. Next, we will introduce our resampling criteria.

\begin{algorithm}[tb]
   \caption{recover($x$,$G$,$R_i$)}
   \label{alg:recovery}
\begin{algorithmic}
   \STATE {\bfseries Input:} image $x$, generator $G$, criteria $R_i$
   \STATE $z_i^{(0)} \sim \mathcal{N}(0,I)$
   \FOR{$t=1$ {\bfseries to} numiter}
     \STATE $z^{(t)} = z^{(t-1)} - \eta^{(t)} \nabla_z ||x - G(z^{(t-1)})||_2^2 $
     \FOR{$i=1$ {\bfseries to} $d$}
       \STATE $thresh \sim \mathcal{U}(0,1) $
       \IF{$P(R_i^{(t)} | z_i^{(t)}) > thresh$} 
         \STATE $z_i^{(t)} \sim \mathcal{N}(0,1)$
       \ENDIF
     \ENDFOR
   \ENDFOR
\end{algorithmic}
\end{algorithm}

\subsubsection{Hard Cutoff Criteria}
The most straightforward criteria is to set a hard cutoff $c$, where we resample any $z_i$ if $|z_i| > c$. In other words, we resample $z_{i}$ if it is more than $c$ away from the mean of the distribution.
$c$ is a hyperparameter to be tuned. Thus the probability that we want to resample $z_i$ from its original
location determined by gradient descent is
$$ P(R_i | z_i) = [\![ |z_i| < c ]\!] $$
using Iverson bracket notation.

\subsubsection{Logistic Cutoff Criteria}
While the hard cutoff criteria assumes that it is completely impossible for a point to be beyond a certain threshold, we may instead want to loosen this condition and say it is simply less probable for a point to be beyond a certain threshold. One way to
model this probability is using a Logistic function (denoted by $\sigma$) as defined:
$$\sigma(x) = \frac{1}{1+e^{-a(x-b)}}$$
In this function, the hyper-parameter $a$ denotes the steepness of our function, and the hyper-parameter $b$ denotes the midpoint of our function, where $ \sigma (b) = \frac{1}{2}$. We note that the Gaussian distribution from which $z_{i}$ is sampled from is symmetric about 0, while $\sigma(x)$ is not. We can impose symmetry using $\sigma(x)$ by considering the absolute value of the input $x$, making sigma a function of how far $x$ is from 0. In the task at hand, we can thus consider $|z_{i}|$ as the input to the $\sigma(x)$. Using this idea, we get: 
$$ P(R_i | z_i) = \frac{1}{1+e^{-a(|z_i|-b)}} $$
We can interpret these parameters as
$b$ being the value of $z_i$ where we are indifferent between using the value recovered from gradient descent
and resampling, and $a$ being the ``softness'' of the cutoff. Refer to \textit{Figure 2.} for the graphical representation of the probability mapping function.

\begin{figure}[h]
\centering
  \begin{minipage}[h]{0.4\textwidth}
  \centering
    \includegraphics[width=\textwidth]{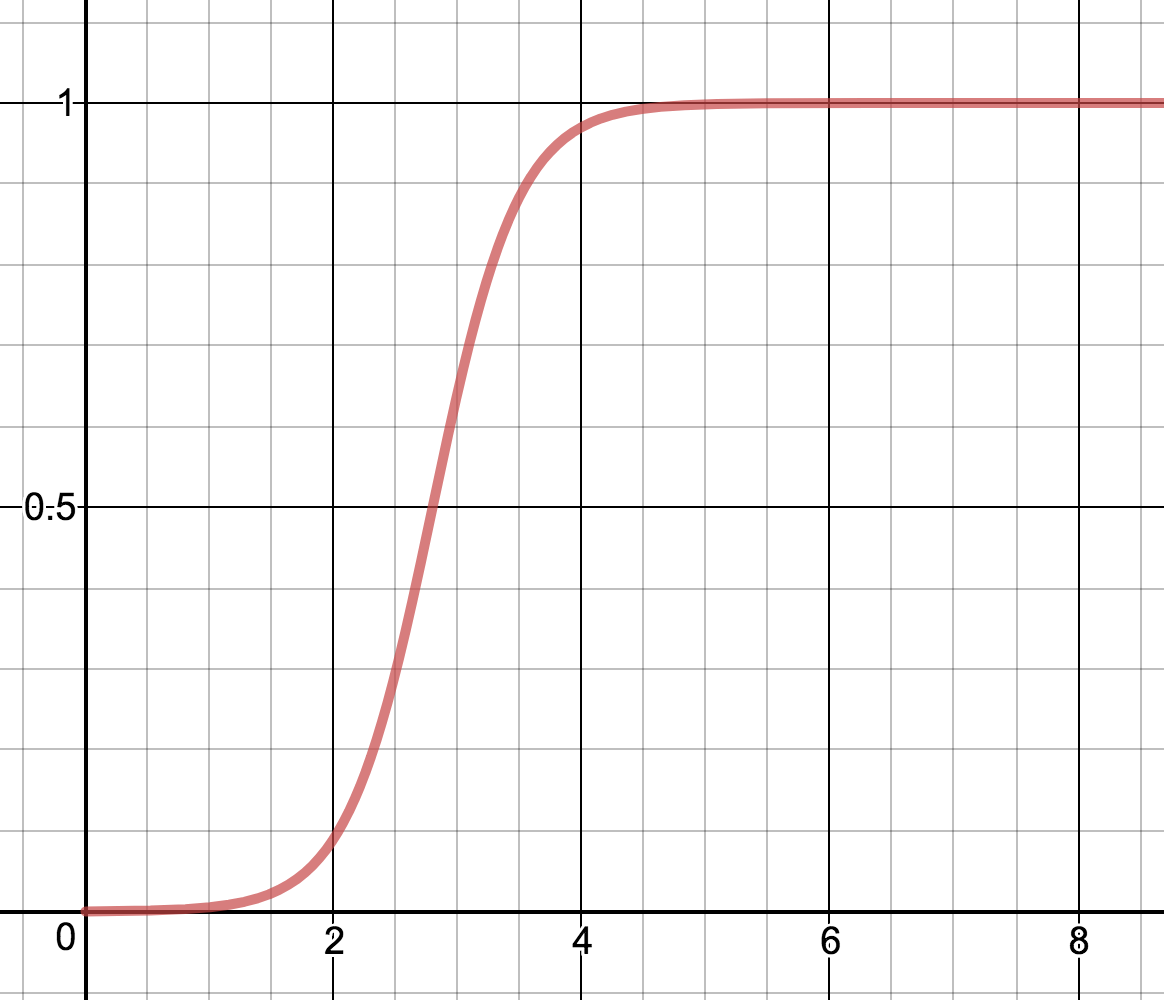}
    \caption{Logistic Criteria for Resampling. This plot shows the quantity $P(R_i | z_i)$ (y-axis) as a function of $|z_{i}|$ (x-axis) }
  \end{minipage}
\end{figure}

\subsubsection{Truncated Normal Cutoff Criteria}
Finally, our last re-sampling criteria is based on the prior knowledge that the latent variable $z$ is sampled from a multivariate Gaussian, implying that each coordinate $z_{i}$ is sampled from a unit Gaussian. Specifically we denote the standard Gaussian-PDF $N(x)$ for each $z_{i}$ as follows:
$$N(x) = \frac{1}{\sqrt{2\pi}} \exp\left\{\frac{-x^{2}}{2}\right\}$$
Intuitively, we want a higher probability of resampling $z_{i}$ when $z_{i}$ adopts a value that occurs with a low probability. Thus, we want the resample probability to be inversely proportional to the probability that we generate $z_{i}$ from the underlying Gaussian distribution. Namely, we want:
$$P(R_i | z_i) \propto \frac{1}{N(z_{i})} \propto \exp \left\{\frac{z_{i}^{2}}{2}\right\}$$
However, we notice that the probability function as given is unbounded with respect to increasing $|z_{i}|$. We deal with this condition by selecting $a$ as a thresholding hyper-parameter. Mathematically, we use $a$ in the following manner:
\begin{equation*}
	P(R_i | z_i) \propto
	\begin{cases}
    	\frac{1}{N(z_{i})},& \text{if } |z_{i}| \leq a\\
        1, & \text{o.w.}
    \end{cases}
\end{equation*}
Solving for the appropriate scaling constant so that $P(R_i | z_i)$ is continuous at $z_{i} = a$, we get the following equation:
\begin{equation*}
	\begin{split}
    	\lim_{z_{i} \to a^{+}} P(R_i | z_i) &= \lim_{z_{i} \to a^{-}} P(R_i | z_i) \\
        \frac{k}{N(a)} & = 1 \\
        k &= N(a)
    \end{split}
\end{equation*}
Thus, we get the following:
\begin{equation*}
	P(R_i | z_i) =
	\begin{cases}
    	\frac{N(a)}{N(z_{i})},& \text{if } |z_{i}| \leq a\\
        1, & \text{o.w.}
    \end{cases}
\end{equation*}

We may interpret the hyper-parameter $a$ as a confidence upper limit for the value that we expect any $z_{i}$ to assume. In other words, any $z_{i}$ that is beyond the $a$-bound is deemed to be too improbable to be generated. See \textit{Figure 3.} for the graphical representation of the probability mapping function.

\begin{figure}[t]
\centering
  \begin{minipage}[t]{0.4\textwidth}
  \centering
    \includegraphics[width=\textwidth]{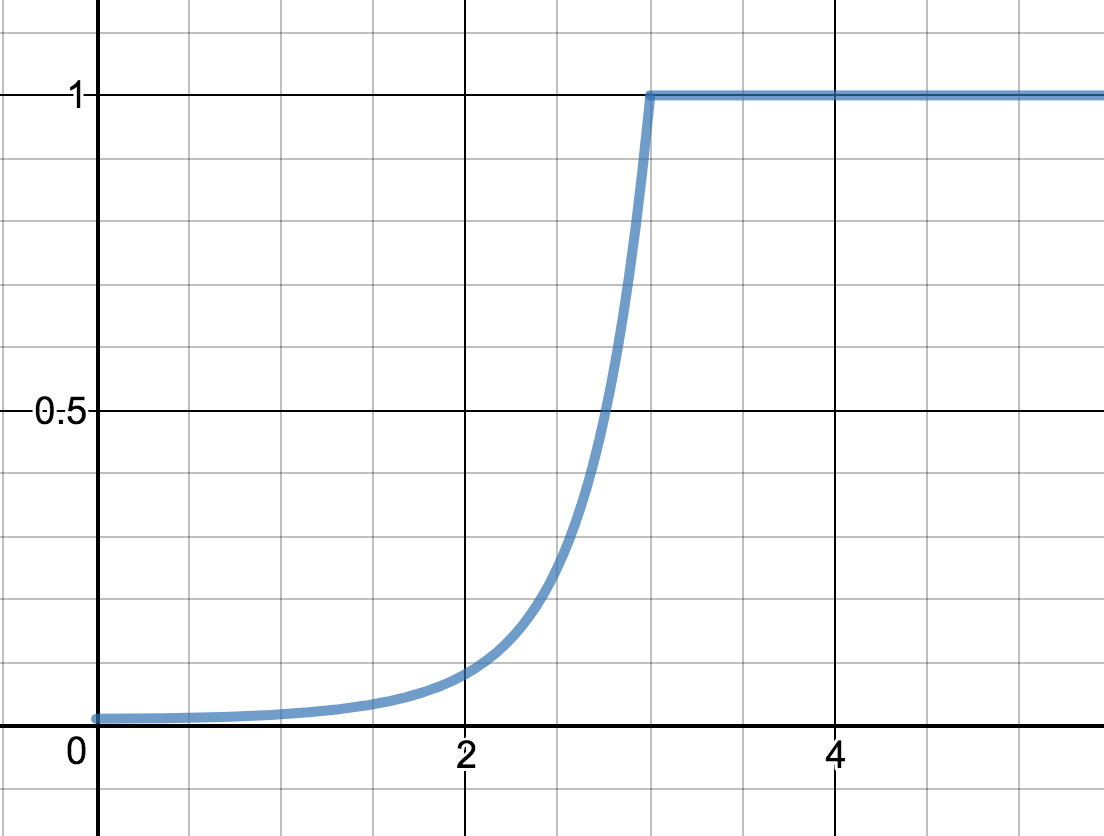}
    \caption{Truncated Normal Criteria for Resampling. This plot shows the quantity $P(R_i | z_i)$ (y-axis) as a function of $|z_{i}|$ (x-axis)}
  \end{minipage}
\end{figure}

\section{Experiments}
\subsection{GAN Modeling and Results}
\subsubsection{Setup}
We utilize the DCGAN framework to create a generative model that generates food images. The model is trained on food images extracted from the Yelp data set (https://www.yelp.com/dataset/).

A fair amount of pre-processing was conducted on the Yelp photos data set. Within the photos data set, we are able to filter out images of food based on a \textit{label} field that can assume a value \textit{food} (this is located in the \textit{photos.json} file of the Yelp data set). At the end of this step, we are left with 121267 available data points to train the GAN. The remaining photos have a great deal of variety in terms of size, resolution, and content (images may contain multiple instances of food and food may not be centered). We further pre-process the data by scaling it to size 64x64 pixels, applying center-cropping, and normalizing the pixel RGB values to be centered around (0.5,0.5,0.5) with standard deviation (0.5,0.5,0.5). To train the models, we use the ADAM optimizer with $2e-4$ as the learning rate and use batch sizes of $64$. We train for 50 epochs, iterating through the entire training set over each epoch. 

\subsubsection{Basic DCGAN}
We use a DCGAN architecture very similar to the one seen in \textit{Figure 1.} as the first model architecture. Note that the output size for the Generator and input size for the Discriminator is 3x64x64 corresponding to the RGB values for each pixel in the image. We base our implementation off of the Pytorch DCGAN example implementation (https://github.com/pytorch/examples/tree/master/ dcgan). The architecture contains two main components and is as follows:\\

\textbf{Generator G:}
\begin{itemize}
    \setlength\itemsep{0.1mm}
    \item Input Layer: Input vector $z$ (length $100$)
    \item Hidden Layer 1: ConvTranspose2D(512x4x4) $\rightarrow$ BatchNorm2D $\rightarrow$ ReLU 
    \item Hidden Layer 2: ConvTranspose2D(256x8x8) $\rightarrow$ BatchNorm2D $\rightarrow$ ReLU  
    \item Hidden Layer 3: ConvTranspose2D(128x16x16) $\rightarrow$ BatchNorm2D $\rightarrow$ ReLU 
    \item Hidden Layer 4: ConvTranspose2D(64x32x32) $\rightarrow$ BatchNorm2D $\rightarrow$ ReLU 
    \item Output Layer: ConvTranspose2D(3x64x64) $\rightarrow$ Tanh 
\end{itemize}

\textbf{Discriminator D:}
\begin{itemize}
    \setlength\itemsep{0.1mm}
    \item Input Layer: Input Image $x$ (3x64x64)
    \item Hidden Layer 1: Conv2D(64x32x32) $\rightarrow$ LeakyReLU(0.2)
    \item Hidden Layer 2: Conv2D(128x16x16) $\rightarrow$ BatchNorm2D $\rightarrow$ LeakyReLU(0.2)
    \item Hidden Layer 3: Conv2D(256x8x8) $\rightarrow$ BatchNorm2D $\rightarrow$ LeakyReLU(0.2)
    \item Hidden Layer 4: Conv2D(512x4x4) $\rightarrow$ BatchNorm2D $\rightarrow$ LeakyReLU(0.2)
    \item Output Layer: Conv2D(1x1x1) $\rightarrow$ Sigmoid
\end{itemize}

See \textit{Figure 4.} for example images generated from the trained basic DCGAN model.

\begin{figure}[h]
\centering
  \begin{minipage}[h]{0.40\textwidth}
  \centering
    \includegraphics[width=\textwidth]{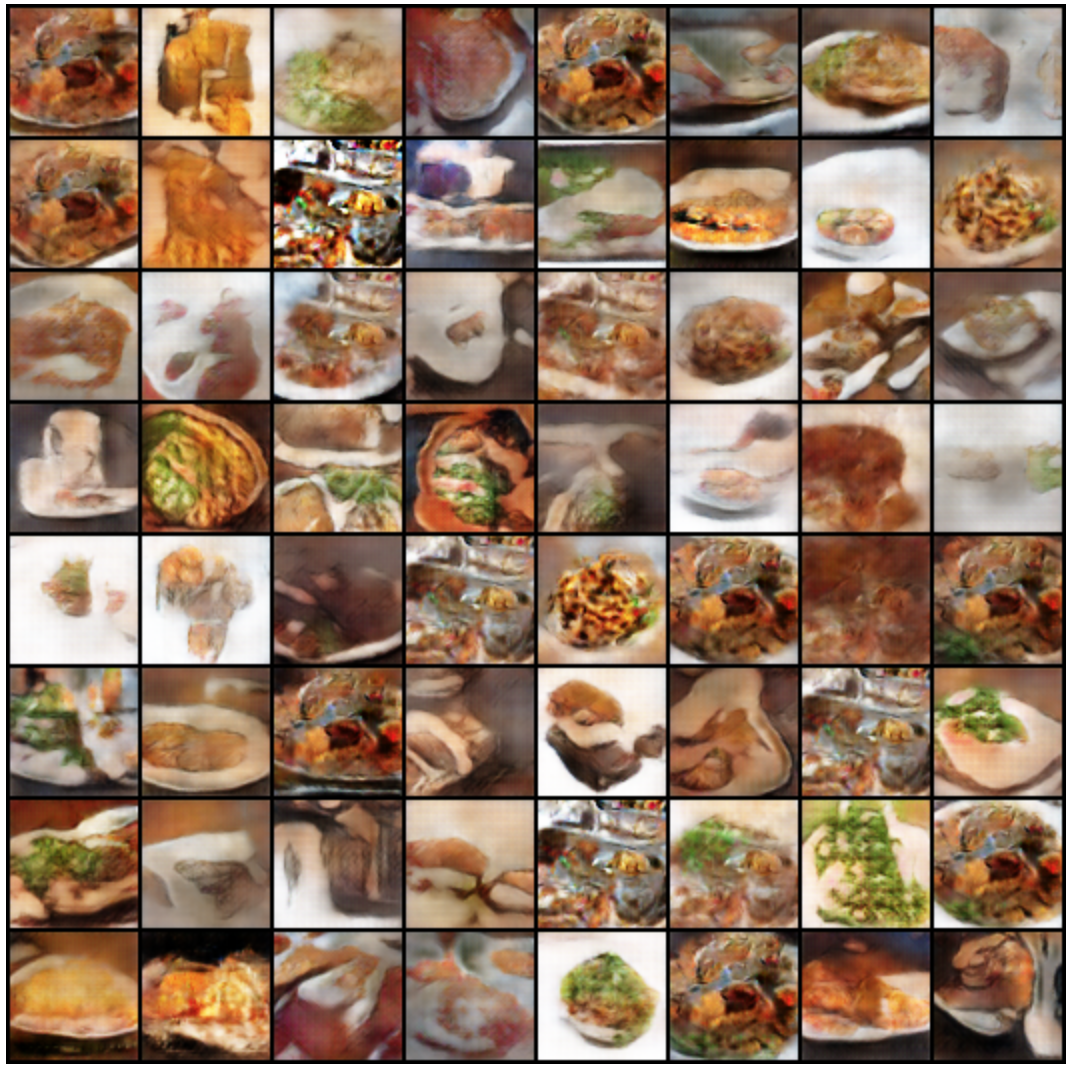}
    \caption{Basic DCGAN model. These are images generated on the $50^{th}$ epoch.}
  \end{minipage}
\end{figure}

\subsubsection{Pixel-Shuffle DCGAN}
Pixel-shuffle is an innovative alternative to transposed convolution for the task of upsampling. Shi et al. \cite{Shi16} notes that using regular transposed convolution requires adding zeros to upscale the input image and that these zeros have to later be filled with meaningful values in a somewhat arbitrary way. Furthermore, a considerably worse point is that these zero values have no gradient information which can be used for backpropagation during training. Pixel shuffling addresses this issue by using regular convolutional layers followed by an image reshaping technique known as a phase shift. Refer to Shi et al.'s paper \cite{Shi16} for more information, and see \textit{Figure 5.} for details.

\begin{figure}[b]
\centering
  \begin{minipage}[b]{0.49\textwidth}
  \centering
    \includegraphics[width=\textwidth]{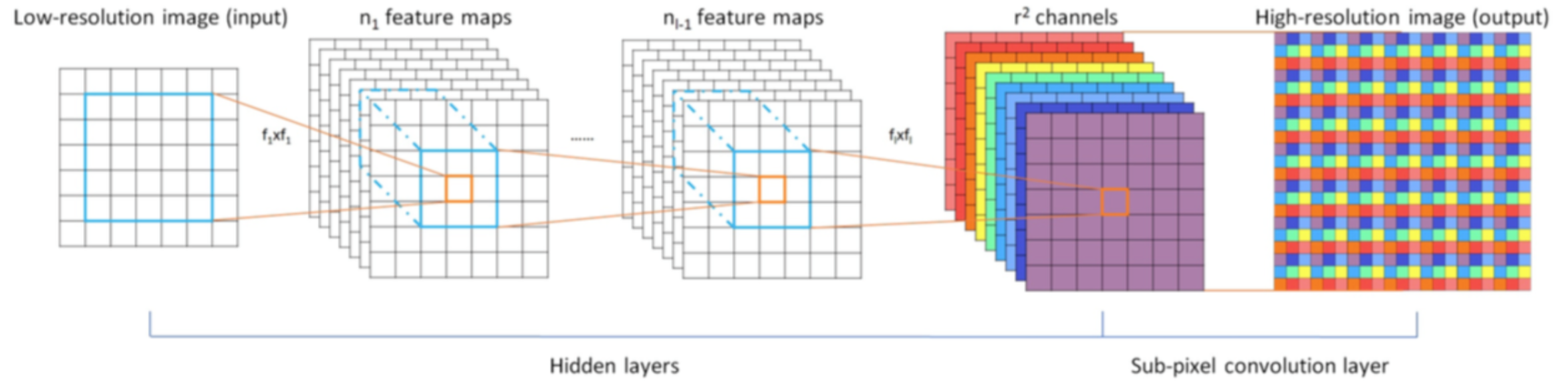}
    \caption{Pixel Shuffle Diagram.}
  \end{minipage}
\end{figure}
Pixel-shuffle is primarily applied to the Generator. In particular, we replace the ConvTranspose2D layer in Hidden Layer 4 with a Pixel-Shuffle layer with upscale parameter 2. Finally, we must accordingly modify the sizes of the previous convolutional layers (HL1, HL2, HL3) to account for the next Pixel-Shuffle layer.. See \textit{Figure 6.} for this model's outputs.
\begin{figure}[h]
\centering
  \begin{minipage}[h]{0.40\textwidth}
  \centering
    \includegraphics[width=\textwidth]{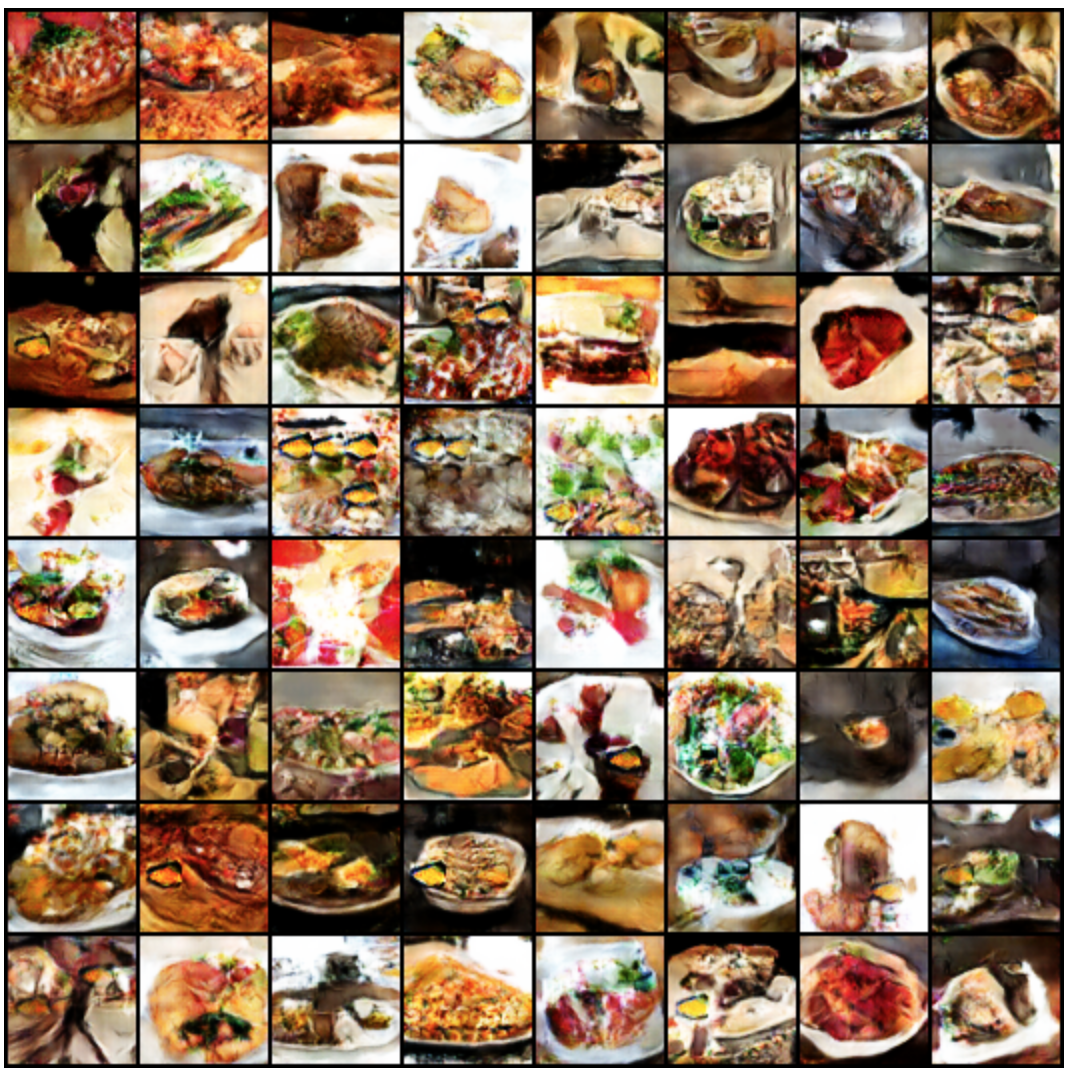}
    \caption{Pixel Shuffle DCGAN model. These are images generated on the $50^{th}$ epoch.}
  \end{minipage}
\end{figure}

\subsubsection{Soft-Labels DCGAN}
Based on the work of Salimans et al. \cite{Salimans16}, label smoothing is shown to reduce the vulnerability of GANs to adversarial examples. To implement this idea, we replace the labels, fake = 0 and real = 1, with labels sampled from uniform distributions over fake = [0,0.3] and real = [0.7,1.2] respectively. \textit{Figure 7.} shows the output of this model.

\begin{figure}[h]
\centering
  \begin{minipage}[h]{0.40\textwidth}
  \centering
    \includegraphics[width=\textwidth]{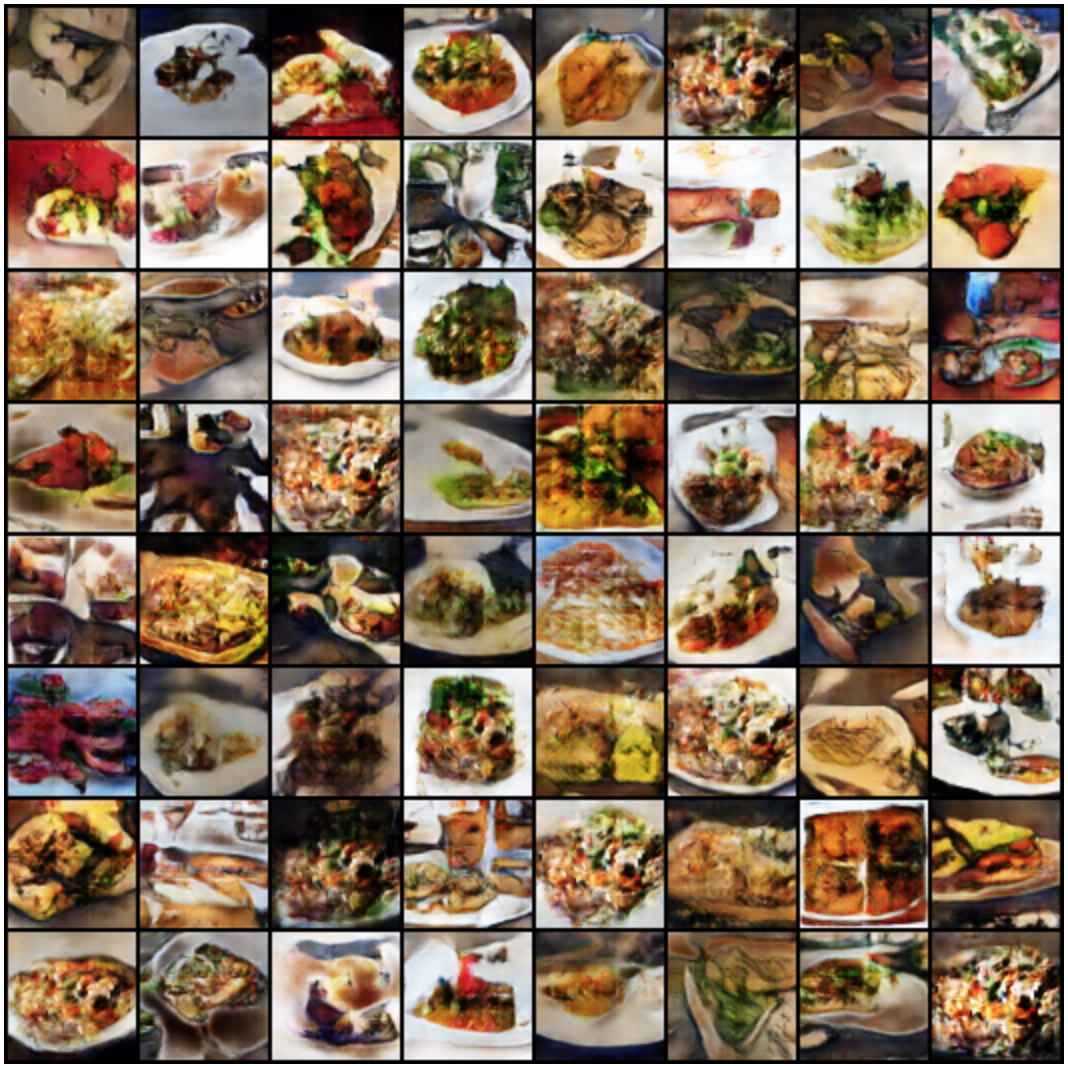}
    \caption{Soft Labels DCGAN model. These are images generated on the $50^{th}$ epoch.}
  \end{minipage}
\end{figure}

\subsection{Resampling Criterion Evaluation}
\subsubsection{Setup}
We used the following experimental setup to evaluate a resampling criterion $R_i$ given our trained GAN $G$ with a multivariate unit normal prior.

\begin{enumerate}
  \setlength\itemsep{0.1mm}
  \item Set $z_{true} \sim \mathcal{N}(0, I)$
  \item Set $x = G(z_{true})$
  \item Get $z_{approx} = $ recover($x$, $G$, $R_i$) where recover() is our latent variable recovery algorithm (Algo. 1)
  \item Calculate reconstruction error as:  $$E(z_{approx}) = ||z_{true} - z_{approx}||_2^2/|z|$$
  where $|z|$ denotes the length of vector $z$.
\end{enumerate}

In our setup, we ran SGD in recover() for 20,000 iterations. The criterion we tested were
\begin{itemize}
  \setlength{\itemsep}{0pt}
  \item disabled (no resampling)
  \item hard($c$) for $c \in \{2.5, 3, 3.5\} $
  \item logistic($a, b$) for $(a,b) \in \{2, 3, 4\} \times \{2, 2.5, 3\} $
  \item trunc\_normal($a$) for $a \in \{2.5, 2.75, 3, 3.25, 3.5\} $
\end{itemize}

To compare these criteria, we fixed 100 values of $z_{true}$ and found the reconstruction error resulting from using each criterion as $R_i$. For each value of $z_{true}$, we chose a random seed, and reseeded the CPU and GPU at the beginning of the recovery algorithm for each criterion. These steps were taken to reduce the chances of any particular run being luckier than any others, but this is still a possibility given that GPU computations are nondeterministic.

Our experiment was run on an Amazon AWS p2.xlarge instance, which has an NVIDIA K80 GPU. We used our pre-trained GAN from section 4.1.4 with soft labels and no pixel shuffle, trained for 20 epochs. Our code is structured like the example PyTorch code from Lipton et al. \cite{Lipton17}, with significant overhauls to set up our experiments. Recovering a single $z$ vector for all 20,000 iterations of SGD takes 70 to 75 seconds in this setup, depending on the resampling criterion being used.

\subsubsection{Results}
Our results are summarized in \textit{Table 1}. The first column is the criterion evaluated with the criterion parameters in parenthesis as denoted above. The next five columns are the percentage of recoveries where the reconstruction error is less than $\epsilon$, for five different values of $\epsilon$. The next column, ``wins,'' gives the percentage of times that the given criterion had lower reconstruction error than the disabled criterion. The next column, ``sig wins,'' gives the percentage of significant wins, which is the percentage of the time that the criterion is better than the disabled criterion by at least a factor of 2, over the amount of times that the criterion differed from the disabled criterion by a factor of 2. The final column is the average reconstruction error for the given criterion.

The reason we kept track of ``significant wins'' is because it is often the case that no resampling or very minor resampling happens during a run of the reconstruction algorithm, and during these runs
the winner between disabled and a given criterion is arbitrary. Due to the same starting place, seed,
and number of iterations, the final reconstruction error should be the same, but the non-determinism of
GPU calculations generally pushes one slightly above the other. Including these runs in the win calculations would even out the numbers between disabled and a given criterion, so we kept track of the number of wins when the difference was significant for evaluation purposes.

\begin{table*}[h!]
\centering
\begin{tabular}{l | r r r r r | r r | r}
  \hline
  \textbf{criterion} & $10^{-4}$ &$10^{-3}$&$10^{-2}$ & $10^{-1}$ & $10^{0}$&wins&sig wins& avg err \\
  \hline
  \hline
  disabled & 0 &  0 &  6 & 41 & 59 & - & - & .864 \\
  \hline
  hard(2.5) & 2 & 10 & 27 & 70 & 80 & 76 & 80.00 & .362 \\
  hard(3) & 3 & 14 & 30 & 62 & 67 & 80 & 80.33 & .577 \\
  hard(3.5) & 0 &  2 & 10 & 58 & 66 & 68 & 76.67 & .671 \\
  \hline
  trunc\_norm(2.5) & 2 &  9 & 38 & 79 & 93 & 82 & 86.75 & .182 \\
  trunc\_norm(2.75) & 2 & 10 & 29 & 66 & 80 & 88 & 96.67 & .408 \\
  trunc\_norm(3) & 4 &  9 & 19 & 63 & 74 & 85 & 93.62 & .516 \\
  trunc\_norm(3.25) & 2 &  7 & 23 & 62 & 73 & 81 & 85.71 & .514 \\
  trunc\_norm(3.75) & 1 &  3 &  8 & 52 & 67 & 64 & 77.50 & .670 \\
  \hline
  logistic(2, 2) & 9 & 19 & 47 & 91 & 91 & 93 & 93.90 & .162 \\
  logistic(3, 2) & 6 & 16 & 48 & 85 & 89 & 90 & 91.25 & .202 \\
  logistic(4, 2) & 4 & 16 & 45 & 82 & 93 & 90 & 93.42 & .183 \\
  logistic(2, 2.5) & 4 & 15 & 41 & 74 & 86 & 89 & 92.11 & .303 \\
  logistic(3, 2.5) & 4 & 13 & 39 & 79 & 86 & 91 & 93.42 & .280 \\
  logistic(4, 2.5) & 2 &  9 & 39 & 77 & 83 & 89 & 94.67 & .328 \\
  logistic(2, 3) & 2 &  5 & 20 & 64 & 74 & 85 & 94.83 & .528 \\
  logistic(3, 3) & 2 &  7 & 23 & 62 & 75 & 80 & 92.16 & .546 \\
  logistic(4, 3) & 1 &  6 & 15 & 55 & 69 & 77 & 87.50 & .644 \\
  \hline
\end{tabular}
\caption{Results from the resampling criteria experiment. See section 4.2.2 for an explanation.}
\end{table*}

\subsubsection{Analysis}
There are many evaluation metrics one can use to judge relative performance of these criteria. 
Depending on the use case of this algorithm, here are some to consider:
\begin{itemize}
  \setlength{\itemsep}{0pt}
  \item \textbf{Average Error:} The simplest way to compare these criteria is to choose the one with the smallest reconstruction error averaged across the 100 runs. The winner here is logistic(2,2), with trunc\_norm(2.5) and logistic(4,2) as runner ups.
  \item \textbf{Improvement:} Another way to compare these criteria is to see how well they improve over the baseline of resampling disabled. The criterion with the most wins over disabled is logistic(2,2), followed by logistic(3, 2.5).
  \item \textbf{Significant Improvement:} If we only care about significant runs, where a criterion differs with the baseline by a factor of two, then we could compare criteria based on significant wins. The winner here is trunc\_norm(2.75), followed by logistic(2,3).
  \item \textbf{Reliability:} If we want to have the algorithm be reliable, meaning that it produces reasonable reconstructions most of the time, then we should expect it to reconstruct $z$ within a threshold of $10^0$ for the highest proportion of runs. trunc\_norm(2.5) and logistic(4,2) are tied in this regard with 93 each, and to break this tie we can compare the proportion of runs where they reconstruct $z$ within a threshold of $10^{-1}$, which makes logistic(4,2) the winner. The reason we go from higher thresholds to lower thresholds in this comparison is because we are preferring a reconstruction that does a decently good job most of the time over a reconstruction that gets lucky some times and performs poorly other times.
\end{itemize}

Across the board for all of our metrics, all of our resampling criteria outperform the baseline algorithm (the one with resampling disabled). Our truncated normal and logistic criteria, as expected, outperform the hard cutoff with optimal parameters set. In general, it appears that the more aggressive of our parameter settings (like hard(2.5), trunc\_norm(2.5), and logistic(4,2)) score better than the more conservative parameter settings. This suggests that the probability that SGD correctly moved a coordinate of $z$ to a magnitude greater than 2.5 is fairly unlikely.

None of our resampling criteria were able to reproduce the results of Lipton et al. \cite{Lipton17}, where they successfully recovered all tested latent vectors with a reconstruction error below $10^{-4}$. Their GAN was trained with a uniform prior, and it appears that reconstructing latent vectors with a uniform prior is a much simpler task: for their runs with clipping (resampling) disabled, all recovered latent vectors had a reconstruction error below $10^{-1}$, while for our runs with resampling disabled, 41\% of recovered latent vectors had a reconstruction error below $10^{-1}$.

\section{Applications}
\subsection{Latent Variable as Embedding}
The latent variable corresponding to a particular image may be used as a feature embedding. In the experiment conducted, we first examined the images generated by the same GAN (soft-labels) from the coordinate unit vectors of the latent space. We define these coordinate unit vectors $\hat e_{i}$'s as follows:
$$\hat e_{i} = \big \langle \mathbbm{1}\{i = 1\}, \mathbbm{1}\{i = 2\}, ..., \mathbbm{1}\{i = 99\}, \mathbbm{1}\{i = 100\} \big \rangle $$
In the GAN we trained, the latent vectors are of size 100 and thus there are 100 different coordinate unit vectors. We then also examine the images generated by the sum of each pair of the unit coordinate vectors. A specific example is demonstrated in \textit{Figure 9}. In this example, the first image (denote as $x_{1}$) on the left hand side was generated from $\hat e_{1}$ and the second image (denoted as $x_{2}$) on the left hand side was generated from $\hat e_{2}$ . 
\begin{figure}[b]
\centering
  \begin{minipage}[b]{0.49\textwidth}
  \centering
    \includegraphics[width=\textwidth]{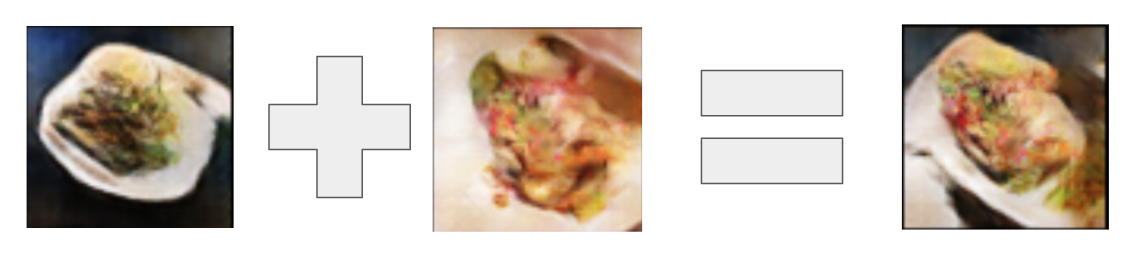}
    \caption{Example of how latent variable can be used as an embedding. For the soft-label GAN, this example shows $x_{1} = G(\hat e_{1})$, $x_{2} = G(\hat e_{2})$, and $x_{o} = G(\hat e_{1}+\hat e_{2})$}.
  \end{minipage}
\end{figure}
Finally, the image on the right hand side (denoted as $x_{o}$)was generated from $\hat e_{1} +\hat e_{2}$. This example is significant because it seems that $x_{o}$ is simply an image where the contents of $x_{2}$ is placed onto the plate in $x_{1}$ over the food contents of $x_{1}$. This embedding interpretation of the latent variable is significant because the embedding can be used as a tool for dimensionality reduction and feature selection. Furthermore, although this embedding may change drastically for different data sets and different types of GAN architecture, it is ultimately learned through a data motivated method and may yield some deeper insight into the underlying data.

\subsection{Latent Vector Interpolation}
One common method for visualizing the latent space of a generative model is interpolation between points. Essentially, given two points in the latent space, we seek to traverse the path between the two points in small, discrete steps. The purpose of this is to visualize the path by using the generator to map the points along the path in the latent vector space to the space of image data. Thus, our final output would be a sequence of images. 

Our implementation of latent vector interpolation works according to the above process with some extra considerations. First, recall that the prior distribution of our latent $z$ vectors is a multivariate Gaussian with mean 0. Then, White \cite{White16}, has shown that for high-dimensional latent spaces, a linear interpolation (traversing the straight-line path) gives poor results. This is due to the fact that in high-dimensional spaces a linear interpolation will pass through points that are highly unlikely given the prior. Instead, White suggests using a spherical linear interpolation (SLERP) for best results. We use the following formula to calculate the path using SLERP between two points $z_1$ and $z_2$.
\begin{align*}
    \text{Slerp}(z_1, z_2, \mu) = \frac{\sin(1-\mu)\theta}{\sin(\theta)}q_1 + \frac{\sin(\mu \theta)}{\sin(\theta)}z_2
\end{align*}
Note that SLERP is equivalent to traversing the great circle of a hyperdimensional sphere. In \textit{Figure 8.}, we present an example of a spherical interpolation starting from a single latent vector and traversing the entire great circle back to the original vector. As one can see, we achieve nice results with this method, with smooth transitions between images. 

\begin{figure}[h]
\centering
  \begin{minipage}[h]{0.40\textwidth}
  \centering
    \includegraphics[width=\textwidth]{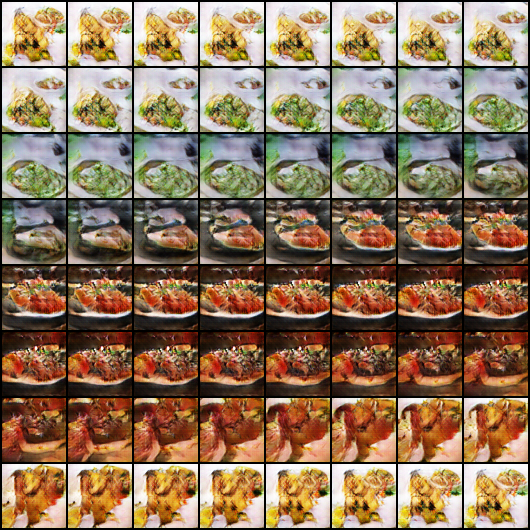}
    \caption{SLERP method for latent vector interpolation}
  \end{minipage}
\end{figure}

\section{Conclusion and Future Work}
We trained a GAN on food photos from the Yelp dataset and through several adjustments to the vanilla GAN formulation, managed to achieve photo-realistic results. One of these adjustments was using a Gaussian prior as opposed to a uniform prior for the latent space, and we sought to improve the current methods of latent variable recovery for latent spaces with Gaussian priors. Our techniques improved upon the currently used methods, and we demonstrated our algorithm's potential applications to unsupervised representation learning.

To build upon this work, we plan on testing our recovery algorithm on different GANs that were trained on different datasets. By using more standard datasets, it would be easier to compare our results to the results presented in other papers. While the logistic(2,2) and trunc\_norm(2.75) criteria performed well on the food GAN, it is quite likely that the optimal parameters of the reconstruction criteria differ across latent spaces. Additionally, we plan on exploring further applications of our GAN's latent variable space to unsupervised learning.

\section*{Contributions} 
Each of the authors participated in the conceptualization and research aspect of the final project. We each read and shared a variety of different related papers regarding the topics of GANs, GAN training, and latent vector recovery. With regards to implementation, Nicholas Egan integrated the basic DCGAN code, experimented with and trained the soft-label DCGAN model, wrote implemented the latent vector recovery algorithm, and implemented the logistic resampling criteria. Jeffrey Zhang experimented with and trained the pixel-shuffle DCGAN and created the truncated normal resampling criteria. Kevin Shen implemented the SLERP method of interpolation and created the hard cutoff resampling criteria. We all contributed to creating the final write up as well as interpretation of the experimental results.

\bibliography{generalized_recovery.bib}
\bibliographystyle{icml2013}

\end{document}